# Predicting Prostate Cancer-Specific Mortality with A.I.-based Gleason Grading

**Authors**:
Ellery Wulczyn[1,†], Kunal Nagpal[1,†], Matthew Symonds[1], Melissa Moran[1], Markus Plass[2], Robert Reihs[2], Farah Nader[2], Fraser Tan[1], Yuannan Cai[1], Trissia Brown[3], Isabelle Flament-Auvigne[3], Mahul B. Amin[4], Martin C. Stumpe[5], Heimo Müller[2], Peter Regitnig[2], Andreas Holzinger[2], Greg S. Corrado[1], Lily H. Peng[1], Po-Hsuan Cameron Chen[1], David F. Steiner[1], Kurt Zatloukal[2], Yun Liu[1,‡], Craig H. Mermel[1,‡]

**Affiliations**:
[1]Google Health, Palo Alto, CA, USA
[2]Medical University of Graz, Graz, Austria
[3]Work done at Google Health via Advanced Clinical, Deerfield, IL, USA
[4]Department of Pathology and Laboratory Medicine, University of Tennessee Health Science Center, Memphis, TN, USA
[5]Work done at Google Health. Present address: Tempus Labs Inc, Chicago, IL, USA
[†]Equal contribution, order determined randomly
[‡]Equal contribution

Address correspondence to: Y.L. (liuyun@google.com), K.Z. (kurt.zatloukal@medunigraz.at)

# Abstract

Gleason grading of prostate cancer is an important prognostic factor but suffers from poor reproducibility, particularly among non-subspecialist pathologists.[1–7] Although artificial intelligence (A.I.) tools have demonstrated Gleason grading on-par with expert pathologists,[8–12] it remains an open question whether A.I. grading translates to better prognostication. In this study, we developed a system to predict prostate-cancer specific mortality via A.I.-based Gleason grading and subsequently evaluated its ability to risk-stratify patients on an independent retrospective cohort of 2,807 prostatectomy cases from a single European center with 5-25 years of follow-up (median: 13, interquartile range 9-17). The A.I.'s risk scores produced a C-index of 0.84 (95%CI 0.80-0.87) for prostate cancer-specific mortality. Upon discretizing these risk scores into risk groups analogous to pathologist Grade Groups (GG), the A.I. had a C-index of 0.82 (95%CI 0.78-0.85). On the subset of cases with a GG in the original pathology report (n=1,517), the A.I.'s C-indices were 0.87 and 0.85 for continuous and discrete grading, respectively, compared to 0.79 (95%CI 0.71-0.86) for GG obtained from the reports. These represent improvements of 0.08 (95%CI 0.01-0.15) and 0.07 (95%CI 0.00-0.14) respectively. Our results suggest that A.I.-based Gleason grading can lead to effective risk-stratification and warrants further evaluation for improving disease management.

# Introduction

Prostate cancer affects 1 in 9 men in their lifetime[13] but disease aggressiveness and prognosis can vary substantially among individuals. The histological growth patterns of the tumor, as characterized by the Gleason grading system, are a major determinant of disease progression and criterion for selection of therapy. Based on the prevalence of these patterns, one of five Grade Groups (GG) is assigned.[14] The GG is among the most important prognostic factors for prostate cancer patients, and is used to help select the treatment plan most appropriate for a patient's risk of disease progression.[15]

The Gleason system is used at distinct points in the clinical management of prostate cancer. For patients undergoing diagnostic biopsies, if tumor is identified, the GG impacts the decision between active surveillance versus definitive treatment options such as surgical removal of the prostate or radiation therapy.[15] For patients who subsequently undergo a surgical resection of the prostate (radical prostatectomy), the GG is one key component of decisions regarding adjuvant treatment such as radiotherapy or hormone therapy.[16,17] In large clinical trials, use of adjuvant therapy following prostatectomy has demonstrated benefits such as improved progression-free survival for some patients, but can also result in substantial adverse side effects[18–20]. As such, several post-prostatectomy nomograms[21] have been developed in order to better predict clinical outcomes following definitive treatment, with the goal of identifying the patients most likely to benefit from adjuvant therapy. Gleason grading of prostatectomy specimens represents a key prognostic element in many of these nomograms and is a central component of the risk categories defined by the National Comprehensive Cancer Network[17].

Due to the complexity and intrinsic subjectivity of the system, Gleason grading suffers from large discordance rates between pathologists (30-50%)[1–6]. However, grades from experts (such as those with several years of experience, primarily practicing urologic pathology or those with urologic subspeciality training) are more consistent and result in more accurate risk stratification than grades from less experienced pathologists[7,22–24], suggesting an opportunity to improve the clinical utility of the system by improving grading consistency and accuracy. To this end, several artificial intelligence (A.I.) algorithms for Gleason grading have been developed and validated using expert-provided Gleason scores.[9–12] However, an evaluation of the prognostic value of these algorithms and a direct comparison to the prognostic value of Gleason grading provided by pathologists has not been conducted. While the GG for biopsies as well as prostatectomy specimens both provide important prognostic information[14], retrospective studies to evaluate long-term clinical outcomes is more straightforward from prostatectomy cases given widely divergent treatment pathways following biopsy alone.

Building on prior work[8,10], we first trained an A.I. system to accurately classify and quantitate Gleason patterns on prostatectomy specimens, and further demonstrate that A.I.-based Gleason pattern quantitations can be used to provide better risk stratification than the Gleason Grade Groups from the original prostatectomy pathology reports.

# Results

All archived slides in prostatectomy cases from 1995-2014 at the Biobank at the Medical University of Graz in Austria[25] were digitized. After excluding 9 cases for death within 30 days of surgery and 8 cases without evidence of prostate cancer in the resection, 2,807 cases remained (Supplementary Figure S1). The median follow-up time was 13.1 years (interquartile range 8.5-17.2). These cases were grouped into two validations sets: all cases (validation set 1) and the subset of cases from 2000-2014 for which Gleason grading was performed at the time of pathologic diagnosis and provided in the final pathology report (n=1,517 cases, validation set 2). Descriptive statistics for both validation sets are provided in Table 1.

For each case, the A.I. algorithm assessed the tumor composition and output percentages for the 3 different Gleason patterns (%GP3, %GP4, %GP5). We fit a Cox proportional hazards regression model directly on these percentages to produce continuous A.I. risk scores, using leave-one-out-cross-validation to "adjust for optimism"[14]. On validation set 1, this continuous A.I. risk score achieved a C-index of 0.84 (95%CI 0.80-0.87) (Table 2). In pre-specified primary analysis, on validation set 2, the C-index for the A.I. risk score (0.87) was significantly greater than the C-index for the GG obtained from the original pathology report (0.79), an improvement of 0.08 (95%CI 0.01-0.15).

To provide an additional comparison to pathologists' GG categorizations, we discretized the A.I. risk scores into five "A.I. risk groups" such that the number of cases per risk group matched the number of cases in the corresponding GG. Similar to the A.I. risk score, the C-index for the A.I. risk groups (0.85) was also greater than the C-index for the pathologist GG (Table 2), an improvement of 0.07 (95%CI 0.00-0.14). Furthermore, Kaplan-Meier analyses showed significant risk stratification across A.I. risk groups across both validation sets (p<0.001 for log-rank test, Figure 1) and univariable Cox regression analyses showed higher hazard ratios for higher A.I. risk groups (Supplementary Table S1).

We also evaluated the prognostic performance of the A.I. in the context of the pathologic T-category. Kaplan-Meier analyses showed significant risk stratification across A.I. risk groups even within groups defined by high and low T-category (p<0.001 for log-rank test, Supplementary Figure S2B). Furthermore, using the A.I. risk groups in a multivariable Cox model that also included T-category gave a C-index that trended higher than using the pathology-report derived Grade Groups (Supplementary Figure S2A).

To better understand discordances between the A.I. risk groups and pathologist GG, we first compared 10-year disease-specific survival rates for cases where the A.I. risk group was higher or lower than the pathologist GG (Supplementary Table S3). Within each pathologist-determined GG, the 10-year survival rates were higher for cases where the A.I. provided a lower risk classification, especially for GG ≥ 3. The survival rates also tended to be lower where the A.I. provided a higher risk classification. Second, risk stratification by the A.I.'s risk groups 1-2 vs. 3-5 remained significant within each pathologist-determined GG (Figure 2). In particular, among patients with pathologist GG 3-5, a sizable subgroup (181 of 436, 42%) were assigned A.I. risk groups of 1-2 and these patients did not experience any disease-specific mortality events (Supplementary Table S3, Figure 2).

Finally, we explored the potential benefit of combining the A.I. system and pathologist grading by evaluating an "ensembling" approach. The arithmetic mean of the A.I. risk group and pathologist-provided GG resulted in a C-index of 0.86 (95%CI 0.80-91) vs. 0.79 for pathologists and 0.85 for the A.I. risk groups (Supplementary Figure S2A). Furthermore, qualitative analysis of algorithm and pathologist discordances suggests several ways in which the algorithmic grading and pathologist grading may be complementary, including consistent grading of regions by the AI which may be variably overgraded by pathologists, or identification of small, high grade regions which may otherwise be missed by pathologists.

# Discussion

In this study, we have validated the ability of a Gleason grading A.I. system to risk-stratify patients using an independent dataset of over 2,800 prostatectomy cases with a median of 13 years of follow-up. The A.I. system demonstrated highly effective risk stratification and, in pre-specified primary analysis, provided significantly better risk stratification than GGs obtained from the original pathology reports.

After prostatectomy, adjuvant radiotherapy for patients with high-risk pathological features has been shown to reduce rates of disease recurrence in multiple clinical trials[18–20], and to improve overall survival in some cohorts[26]. Given their prognostic value, Gleason grades represent a key factor in adjuvant therapy decisions, with NCCN practice guidelines suggesting higher risk patients be considered for adjuvant therapy[15]. However, use of adjuvant radiotherapy can cause adverse effects, contributing to low utilization of this treatment option[27] despite there being a subset of patients who would likely benefit. While risk stratification tools such as nomograms (in which the Gleason Score is among the most prognostic factors)[21] and molecular tests[28] have been developed, selection of patients for adjuvant therapy post prostatectomy remains a difficult task[15]. Given the ability of the A.I. to provide significant risk stratification among patients most likely to consider adjuvant therapy (GG 3-5 and pT3 and above, Supplementary Figure S2C), our results suggest that the A.I. risk-score could be particularly useful for informing adjuvant therapy decisions. Evaluation of whether additional prognostic value can be obtained by combining the A.I. risk score with existing prognostic tools such as nomograms and molecular approaches is also warranted.

The A.I. system may also contribute to clinical decision making by directly assisting pathologist grading as a computer-aided diagnostic (CADx) tool. Prior work has shown that a CADx tool for Gleason grading can improve grading consistency and accuracy by pathologists, with pathologists benefiting from the consistent grading provided by the A.I. while also correcting and overriding unexpected A.I. errors as needed[29,30]. Given the prognostic importance of expertise in pathology review[7], and the scarcity of specialty pathologists in low-income and middle-income countries[31], utilization of the A.I. system as an assistive tool during prostatectomy review has the potential to improve access to consistent, accurate grading, and may ultimately result in grading that more accurately predicts patient outcome.

While not directly comparable due to differences in cohorts and study design, the prognostic performance observed for the pathologist Gleason grading in this cohort is largely consistent with prior work evaluating associations of pathologist grading and clinical outcomes (c-indices of 0.70-0.83 for Grade Groups and biochemical recurrence[14,32,33] and 0.80 for the recent STAR-CAP clinical prognostic grouping and DSS[34]).

Several other works have developed Gleason grading algorithms, though without validating them on clinical outcomes[9,11,35]. Additionally, Yamamoto et al. recently demonstrated the ability to directly learn prognostic histologic features in prostate cancer specimens that correlate with patient outcomes[36]. The present study complements prior work by building upon an extensively validated Gleason system to provide A.I. risk assessments that are directly interpretable by pathologists and utilizing a large independent dataset with long-term clinical follow-up for direct validation of these assessments on patient outcomes.

This study has some limitations. First, without access to treatment information for this cohort, we were unable to evaluate our A.I. within subgroups defined by potentially different treatment pathways following prostatectomy. Next, the Gleason grading system has evolved over the time period in which data was collected for this study, potentially contributing to inconsistencies in grading between pathologists and underestimating the prognostic performance of the GG in the original report. Relatedly, we did not have access to the raw Gleason pattern percentages used by pathologists to determine the Grade Group, which limited comparison with continuous pathologist risk scores. Similarly, the A.I. does not take into account grading subtleties such as grading dominant or codominant nodules, but evaluates the entire case holistically. Next, this study focuses on prostatectomy specimens. The benefit of prostatectomy-based analysis is that the interpretation of prognostication performance in resections is more straightforward than for biopsies due to less divergent post-operative treatment pathways[37]. Future work to validate an accurate A.I. system's prognostic utility on biopsies may provide additional opportunities to inform and improve post-biopsy clinical decisions. Lastly, in addition to Gleason grading, pathologists review cases for additional criteria, including TNM staging, cancer variants[38], and other pathologic findings not evaluated by our system. Therefore, the potential benefits of integrating our A.I. system into a routine pathology workflow will ultimately need to be evaluated in prospective studies.

To conclude, we have validated the ability of an A.I. Gleason grading system to effectively risk-stratify patients on a large retrospective cohort, outperforming the Gleason GG in the original report. We look forward to future research involving the clinical integration and evaluation of the impact of A.I. for improving patient care.

# Methods

## Data

All available slides for archived prostate cancer resection cases between 1995 and 2014 in the Biobank Graz at the Medical University of Graz were retrieved, de-identified, and scanned using a Leica Aperio AT2 scanner at 40X magnification (0.25 µm/pixel). Primary and secondary Gleason patterns (Gleason Scores) were extracted from the original pathology reports, along with pathologic TNM staging, and patient age at diagnosis. Gleason Scores were translated to their corresponding Grade Groups[14]. 22 cases (1%) were indicated as having pathologic T-category in T1 in the original pathology report, which is a categorization reserved for clinical T-category only; these pathology reports were subsequently re-reviewed by a pathologist for appropriate re-categorization. Disease-specific survival (DSS) was inferred from International Classification of Diseases (ICD) codes from the Statistik Austria database. Codes considered for prostate-cancer related death were C61 and C68. Institutional Review Board approval for this retrospective study using anonymized slides and associated pathologic and clinical data was obtained from the Medical University of Graz (Protocol no. 32-026 ex 19/20).

Validation set 1 included all available cases from 1995-2014 after application of the exclusion criteria (n=2,807, Table 1 and Supplementary Figure S1). Because Gleason scoring at the Medical University of Graz was adopted in routine practice from 2000 onwards, validation set 2 included all cases from 2000 onwards for which a Gleason score was available (n=1,517, Table 1). Sensitivity analysis for inclusion of Gleason grades prior to the year 2000 (before Gleason scoring became routine at the institution) is presented in Supplementary Table S4.

All slides underwent manual review by pathologists (See "Pathologist Cohort" in the Supplementary Methods) to confirm stain type and tissue type. Inclusion/exclusion criteria are described in Supplementary Figure S1. Briefly, immunohistochemically stained slides were excluded from analysis and only slides containing primarily prostatic tissue were included. Slides containing exclusively prostatic tissue were included in their entirety. Slides with both prostatic tissue and seminal vesicle tissue were included, but processed using a prostatic tissue model meant to provide only prostatic tissue to the Gleason grading model (see "Prostatic Tissue Segmentation Model" in Supplementary Methods).

## Gleason Grading Model

We previously developed two A.I. systems: one for Gleason grading prostatectomy specimens[8] based on a classic "Inception" neural network architecture, and a second for Gleason grading biopsy specimens based on a customized neural network architecture[10]. For this work, we used the prostatectomy dataset from the first study to train a new model using the customized neural network architecture introduced in the second study. The training dataset contained 112 million pathologist-annotated "image patches" from a completely independent set of prostatectomy cases from the validation data used in this study. Briefly, the system takes as input 512x512 pixel image patches (at 10X magnification, 1 μm per pixel) and classifies each patch as one of four categories: non-tumor, Gleason pattern 3, 4, or 5. The hyperparameters used for training this network were determined using a random grid search over 50 potential settings and are described in Supplementary Table S6.

## A.I. Risk Scores and Risk Groups

The Gleason grading model was run at stride 256 (at 10X magnification, 1 μm per pixel) on all prostate tissue patches. The classification of each patch as non-tumor or GP 3, 4, or 5 was determined via argmax on re-weighted predicted class probabilities[8]. For each case, the percentage of prostate tumor patches that belong to Gleason patterns 3, 4 and 5 were subsequently computed. A.I. risk scores were computed by fitting a Cox regression model using these case-level Gleason pattern percentages as input, and the right-censored outcomes as the events. This approach was pursued first (rather than direct mapping of %GPs to GG as done by pathologists) due to the prognostic importance of precise Gleason pattern quantitation[39], as well as the exhaustive nature of A.I. grading that rarely leads to classifications of GG1 (e.g. 100% GP3) and GG4 (e.g. 100% GP4). Sensitivity analyses evaluating additional ways of obtaining risk groups from %GPs, including direct mapping of %GPs to GG and a temporal-split methodology, demonstrated qualitatively similar results and are presented in Supplementary Table S5.

Gleason pattern 3 percentage was dropped as an input feature to avoid linear dependence between features. Leave-one-case-out cross-validation was used to adjust for optimism, similar to the 10-fold cross validation used in Epstein et al.[14] A.I. risk groups were derived from the A.I. risk scores by discretizing the A.I. risk scores to match the number and frequency of pathologist GG in validation set 2.

## Statistical Analysis

Primary and secondary analyses were prespecified and documented prior to evaluation on the validation sets. The primary analysis consisted of the comparison of c-indices for DSS between pathologist GG and the A.I. risk scores (Table 2). The secondary analysis consisted of the comparison between c-indices for pathologist GG and the discretized A.I. risk groups. All other analyses were exploratory.

The prognostic performance of the pathologist GG, the A.I. risk scores and the A.I. risk groups were measured using Harrel's C-index[40], a generalization of area under the receiver operating characteristic curve (AUC) for time-censored data. Confidence intervals for both the c-index of A.I. and pathologists, and the differences between them, were computed via bootstrap resampling[41] with 1000 samples.

In Kaplan-Meier analysis of the pathologist GG and A.I. risk groups, the multivariate log-rank test was used to test for differences in survival curves across groups. All survival analysis were conducted using the Lifelines python package[42] (version 0.25.4).

## Data availability

This study utilized archived anonymized pathology slides, clinicopathologic variables, and outcomes from the Institute of Pathology and the Biobank at the Medical University of Graz. Interested researchers should contact K. Z. to inquire about access to Biobank Graz data; requests for non-commercial academic use will be considered and require ethics review prior to access.

## Acknowledgments

This work was funded by Google LLC and Verily Life Sciences. The authors would like to acknowledge the Google Health Pathology team for software infrastructure support and data collection. We also appreciate the input of Jacqueline Shreibati, Alvin Rajkomar, and Dale Webster for their feedback on the manuscript. Last but not least, this work would not have been possible without the support of Dr. Christian Guelly and the Biobank Graz, Andrea Berghold, and Andrea Schlemmer from the Institute of Medical Informatics and the efforts of the slide digitization team at the Institute of Pathology.

## Competing interests

E.W., K.N., M.S., M.M., F.T., Y.C., M.C.S., G.S.C., L.H.P., P-H.C.C., D.F.S., Y.L. and C.H.M are current or past employees of Google LLC, own Alphabet stock, and are co-inventors on patents (in various stages) for machine learning using medical images. I.F-A., T.B., and M.B.A. are current or past consultants of Google LLC. M.P., R.R., F.N., H.M., P.R., A.H., and K.Z. are employees of the Medical University of Graz.

## Author contributions

K.N. and M.S. performed the majority of the machine learning development with guidance from P-H.C.C. and Y.L.; K.N., E.W., and P-H.C.C. wrote the technical machine learning software infrastructure. E.W. and K.N. designed the study and pre-registered statistical analyses with input from M.B.A., P-H.C.C., D.F.S, K.Z, Y.L., and C.H.M.. E.W. and K.N. performed statistical analyses. M.M., M.P., and R.R. managed the scanning operations for whole slide image digitization. M.P., R.R., F.N., F.T., A.H., and Y.C. prepared the clinical metadata used in the study. T.B., I.F-A., M.B.A., P.R., and K.Z. provided pathology domain expertise. M.C.S., H.M., G.S.C., L.H.P., K.Z., Y.L., and C.H.M. obtained funding for data collection and analysis, supervised the study, and provided strategic guidance. E.W., K.N., P-H.C.C., D.F.S., and Y.L. prepared the manuscript with input from all authors. E.W. and K.N. contributed equally as co-first authors; Y.L. and C.H.M. contributed equally as co-last authors.

## Code availability

The deep learning framework (TensorFlow) used in this study is available at https://www.tensorflow.org. The deep learning architecture for the Gleason grading model is detailed in prior work[10]. All survival analyses were conducted using Lifelines[42], an open source Python library.

Figures

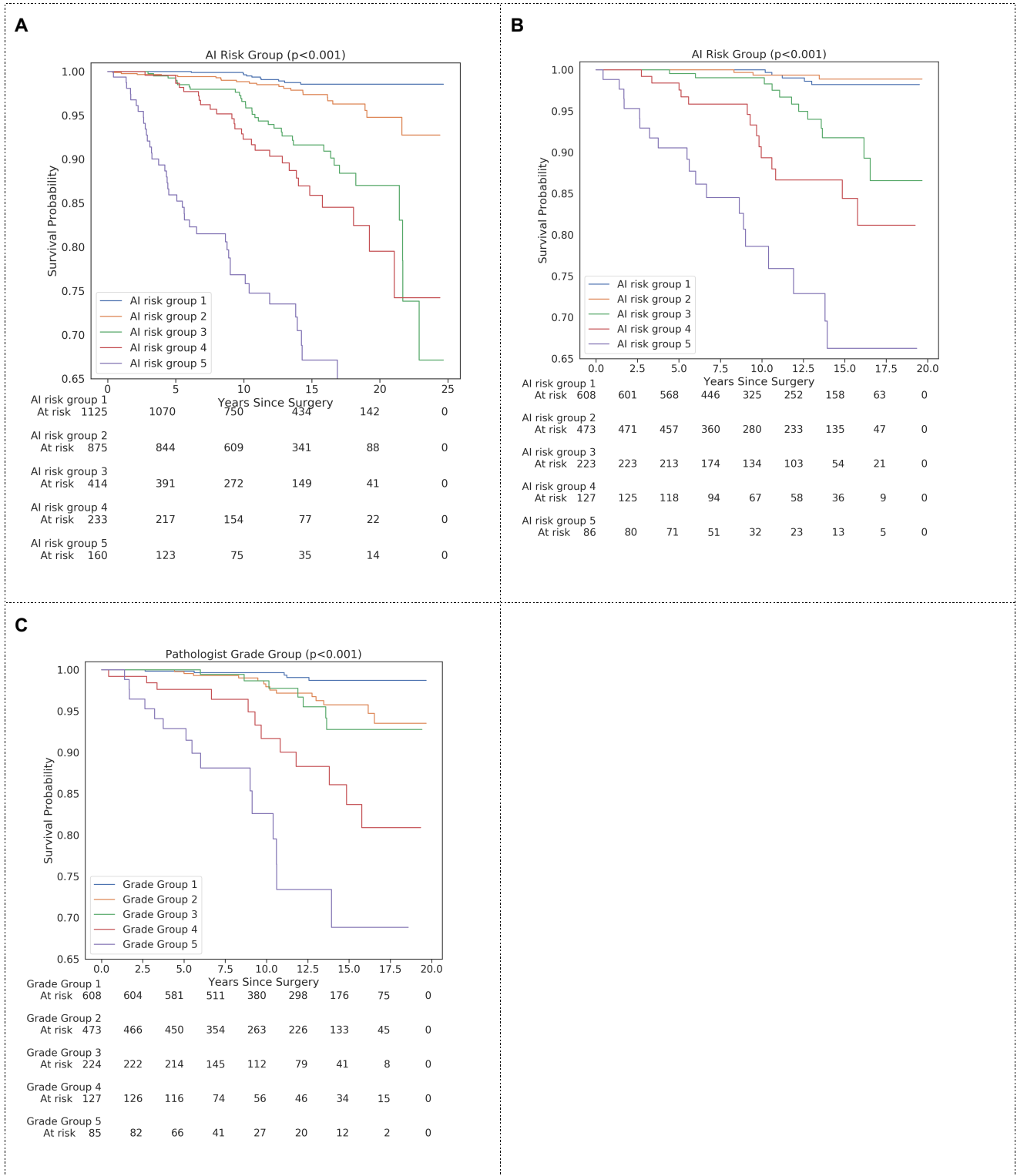

**Figure 1. Kaplan-Meier curves for A.I. and pathologist.** Kaplan-Meier (KM) curves for (A) A.I. risk groups on validation set 1, (B) A.I. risk groups on validation set 2 and (C) pathologist Grade Groups on validation set 2. P-values were calculated using the log-rank test.

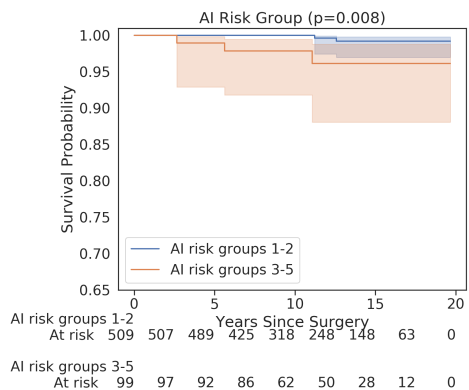
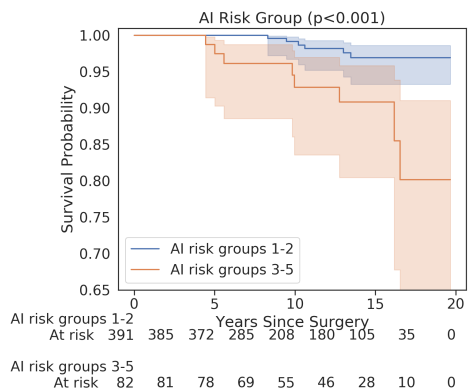
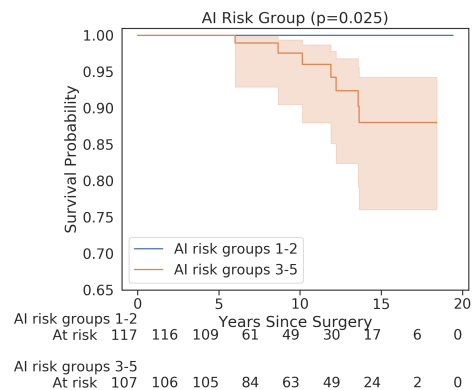
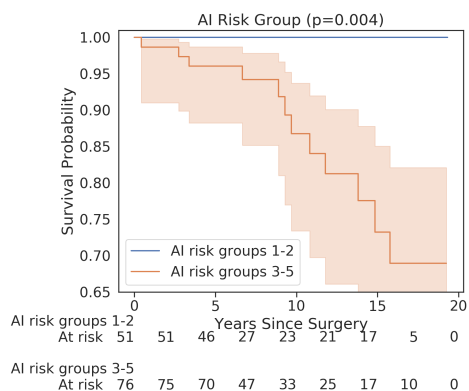
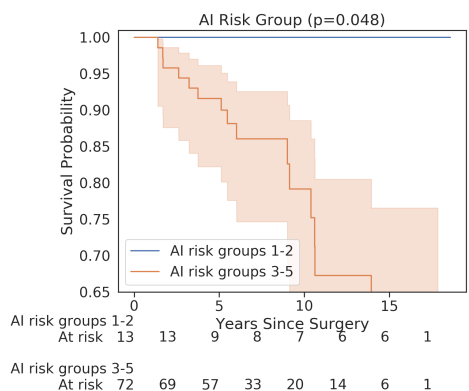

**Figure 2. Sub-stratification of patients by A.I. as Risk Groups 1-2 vs. 3-5 within each pathologist-determined GG.**

# Tables

**Table 1. Cohort characteristics.** Validation set 1 contains all prostatectomy cases from the Biobank Graz between 1995-2014. Validation set 2 is a subset of validation set 1 and contains all prostatectomy cases between 2000-2014 where a diagnostic Grade Group was recorded.

|  | Validation Set 1 | Validation Set 2 |
|---|---|---|
| **Number of cases** | 2,807 | 1,517 |
| **Number of slides** | 83,645 | 47,626 |
| **Overall survival (OS)** | | |
| Median years of follow-up (interquartile range) | 13.1 (8.5, 17.2) | 11.2 (7.4, 15.2) |
| Censored | 2150 | 1306 |
| Observed | 657 | 211 |
| **Disease-specific survival (DSS)** | | |
| Censored | 2,673 | 1,464 |
| Observed | 134 | 53 |
| **Grade Group** | | |
| 1 | 611 | 608 |
| 2 | 476 | 473 |
| 3 | 224 | 224 |
| 4 | 128 | 127 |
| 5 | 85 | 85 |
| Unknown | 1283 | 0 |
| **Pathologic T-stage** | | |
| T2 | 1,640 | 1,113 |
| T3 | 791 | 366 |
| T4 | 25 | 6 |
| Unknown | 351 | 32 |
| **Age at diagnosis** | | |
| <60 | 952 | 537 |
| 60-70 | 1546 | 817 |
| $\geq$70 | 309 | 163 |
| **Margin status** | | |
| Negative | 448 | 153 |
| Positive | 242 | 96 |
| Unknown | 2117 | 1268 |

**Table 2. C-Index for pathologist and A.I. grading.** The A.I. risk score (B) is a continuous risk score from a Cox regression fit on Gleason pattern percentages from the A.I.. The A.I. risk group (C) is a discretized version of the A.I. risk score. The discretization was done to match the number and frequency of pathologist Grade Groups in validation set 2. In validation set 2, the c-index for the A.I. risk score was statistically significantly higher than that for the pathologists' Grade Group ($p<0.05$, prespecified analysis). Bold indicates the highest value in each column (dataset).

|  | C-index [95%CI] | |
| --- | --- | --- |
|  | Validation Set 1 | Validation Set 2 |
| A) Pathologist Grade Groups | N/A* | 0.79 [0.71, 0.86] |
| B) A.I. risk score (continuous) | **0.84 [0.80-0.87]** | **0.87 [0.81, 0.91]** |
| C) A.I. risk groups (discretized) | 0.82 [0.78-0.85] | 0.85 [0.79, 0.90] |

*Not available because pathologist Grade Groups were not available for all cases in validation set 1 due to the earlier time period.

# Supplementary Material

## Supplementary Methods

### Prostatic Tissue Segmentation Model

In order to collect data for model development, pathologists were asked to coarsely outline extraprostatic tissue and seminal vesicle regions across 221 slides from The Cancer Genome Atlas[43] and previously-digitized de-identified slides from the Naval Medical Center San Diego[8]. Extraprostatic tissue and seminal vesicle annotations were combined into a single 'Extraprostatic Tissue' class. An additional 150 de-identified slides were randomly sampled from the Gleason grading dataset (see "Gleason Grading Model" in the Methods), and any benign or Gleason pattern 3, 4, or 5 annotation was considered to be part of the 'Prostatic Tissue' class.

The resulting 371 slides were randomly split into a training and tuning split. A convolutional neural network, using the same architecture, training methodology, and hyperparameter tuning methodology described for the Gleason Grading model, was trained for the binary 'Extraprostatic Tissue' vs. 'Prostatic Tissue' task, with a resulting AUC of 0.99 on the tuning set. The threshold for binarization was chosen to achieve 97% precision (at 84% recall) of prostatic tissue.

### Pathologist Cohort

Manual pathologist reviews for slides' stain and tissue types were performed by a cohort of 19 US board-certified pathologists across 11 states and 2 non-US trained pathologists. The median of years of experience amongst this cohort was 11 (range: 2-25).

# Supplementary Tables

**Supplementary Table S1. Hazard ratios for pathologist Grade Group and A.I. risk group**

Hazard ratios from univariable Cox regression models for pathologist Grade Group and A.I. risk groups. P-values were computed from a Wald test.

|  | Validation Set 1 | | Validation Set 2 | |
| --- | --- | --- | --- | --- |
|  | Hazard ratio [95%CI] | P-value | Hazard ratio [95%CI] | P-value |
| **Pathologist Grade Group** | | | | |
| 1 | * N/A | | 1.0 (reference) | - |
| 2 | | | 3.85 [1.39, 10.70] | p=0.010 |
| 3 | | | 4.68 [1.49, 14.76] | p=0.009 |
| 4 | | | 14.30 [5.03, 40.62] | p<0.001 |
| 5 | | | 35.87 [13.00, 98.97] | p<0.001 |
| **A.I. risk group** | | | | |
| 1 | 1.0 (reference) | - | 1.0 (reference) | - |
| 2 | 2.83 [1.34, 5.98] | p=0.006 | 0.71 [0.17, 2.97] | p=0.641 |
| 3 | 9.55 [4.70, 19.37] | p<0.001 | 6.23 [2.19, 17.69] | p<0.001 |
| 4 | 13.99 [6.77, 28.92] | p<0.001 | 13.16 [4.74, 36.54] | p<0.001 |
| 5 | 39.96 [20.04, 79.69] | p<0.001 | 35.54 [13.26, 95.27] | p<0.001 |

*Not available because pathologist Grade Groups were not available for all cases in validation set 1 due to the earlier time period.

**Supplementary Table S2. Hazard ratios for A.I. Gleason pattern percentages.** Hazard ratios from multivariable Cox regression models on A.I. Gleason pattern percentages. Gleason pattern percentages from pathologists were not available from the clinical reports for these cohorts. Hazard ratios represent the risk increase per 10 percentage point increase in the respective pattern. P-values were computed from a Wald test.

|  | Validation Set 1 | | Validation Set 2 | |
| --- | --- | --- | --- | --- |
|  | Hazard ratio [95%CI] | P-value | Hazard ratio [95%CI] | P-value |
| % Gleason Pattern 3 | 1.0 (reference) | - | 1.0 (reference) | - |
| % Gleason Pattern 4 | 1.48 [1.37, 1.60] | <0.001 | 1.58 [1.39, 1.79] | <0.001 |
| % Gleason Pattern 5 | 1.51 [1.41, 1.61] | <0.001 | 1.63 [1.46, 1.82] | <0.001 |

**Supplementary Table S3. 10-year disease-specific survival rates for disagreements between pathologist and A.I..** Kaplan-Meier estimates of 10-year survival rates for validation set 2 cases. For each pathologist Grade Group (GG) survival rates are shown for all cases ("All"), cases where the A.I. risk group was lower than the pathologist GG ("Lower"), cases where the A.I. risk group was the same as the pathologist GG ("Same") and cases where the A.I. risk group was higher than the pathologist GG ("Higher"). Numbers in square braces indicate 95% confidence intervals, with n indicating the size of the group.

| Pathologist Grade Group | All | A.I. risk group | | |
|---|---|---|---|---|
| | | Lower | Same | Higher |
| **Grade Group 1** | 1.00 [0.99, 1.00] n=608 | N/A | 1.00 [1.00, 1.00] n=327 | 0.99 [0.97, 1.00] n=281 |
| **Grade Group 2** | 0.98 [0.96, 0.99] n=473 | 1.00 [1.00, 1.00] n=212 | 0.98 [0.94, 1.00] n=179 | 0.93 [0.84, 0.97] n=82 |
| **Grade Group 3** | 0.99 [0.95, 1.00] n=224 | 1.00 [1.00, 1.00] n=117 | 0.98 [0.87, 1.00] n=56 | 0.97 [0.81, 1.00] n=51 |
| **Grade Group 4** | 0.92 [0.83, 0.96] n=127 | 1.00 [1.00, 1.00] n=83 | 0.81 [0.56, 0.93] n=26 | 0.72 [0.33, 0.91] n=18 |
| **Grade Group 5** | 0.83 [0.70, 0.90] n=85 | 0.93 [0.74, 0.98] n=53 | 0.65 [0.42, 0.81] n=32 | N/A |

**Supplementary Table S4. Sensitivity analysis for years included in validation set 2.** All results represent C-index with 95% confidence intervals in square braces. The middle column presents the original validation set 2 analysis comprising all cases with a Gleason score from 2000 onwards; the rightmost column presents analysis comprising all cases with a Gleason score (including those before 2000).

| Year of analysis | 2000-2014 | 1995-2014 |
| --- | --- | --- |
| No. of cases | 1,517 | 1,524 |
| Pathologist Grade Groups | 0.79 [0.71, 0.86] | 0.78 [0.71, 0.85] |
| A.I. risk score | 0.87 [0.81, 0.91] | 0.86 [0.81, 0.91] |
| A.I. risk group | 0.85 [0.79, 0.90] | 0.85 [0.80, 0.90] |

**Supplementary Table S5. Sensitivity analysis for discretization method.** Sensitivity analysis evaluating different ways of obtaining discrete A.I. risk groups from A.I. Gleason pattern percentages. A) Risk scores from a Cox regression model fit on A.I. Gleason pattern percentages were generated via leave-one-out cross-validation (LOOCV) and discretized to match the pathologist Grade Group distribution in validation set 2. B) Cases from 1995-2000 were used to train a Cox regression model on A.I. Gleason pattern percentages. Risk scores from this model on validation set 2 were discretized to match the pathologist Grade Group distribution in validation set 2. C) A.I. Gleason pattern percentages were mapped to discrete risk groups using the same rule-based mapping used by pathologists to determine the Grade Group from Gleason pattern percentages[44].

|  | C-index [95%CI] | |
| --- | --- | --- |
|  | Validation Set 1 | Validation Set 2 |
| A) LOOCV | 0.82 [0.78, 0.85] | 0.85 [0.79, 0.90] |
| B) Temporal split | N/A | 0.86 [0.80, 0.90] |
| C) Rule-based | 0.80 [0.75, 0.84] | 0.84 [0.78, 0.88] |

**Supplementary Table S6. Hyperparameters for training Gleason grading model**

|  | **Gleason Grading Model** | **Prostatic Tumor Segmentation** |
|---|---|---|
| Architecture | Custom TuNAS Architecture[10]<br>L2 Weight Decay: 0.004 | |
| Color perturbations | Saturation delta: 0.80<br>Brightness delta: 0.96<br>Contrast delta: 0.17<br>Hue delta: 0.02 | |
| Learning rate schedule | Exponential decay schedule<br>Base rate: 0.0042<br>Decay rate: 0.95<br>Decay steps: 51,733 steps | Exponential decay schedule<br>Base rate: 0.0001<br>Decay rate: 0.90<br>Decay steps: 25,000 steps |
| RMSProp optimizer | Decay: 0.95<br>Momentum: 0.7<br>Epsilon: 0.001 | Decay: 0.95<br>Momentum: 0.7<br>Epsilon: 0.001 |
| Other | Image input magnification: 10X (1 µm/pixel)<br>Loss function: softmax cross-entropy<br>Batch size: 32 | Image input magnification: 5X (2 µm/pixel)<br>Loss function: softmax cross-entropy<br>Batch size: 16 |

# Supplementary Figures

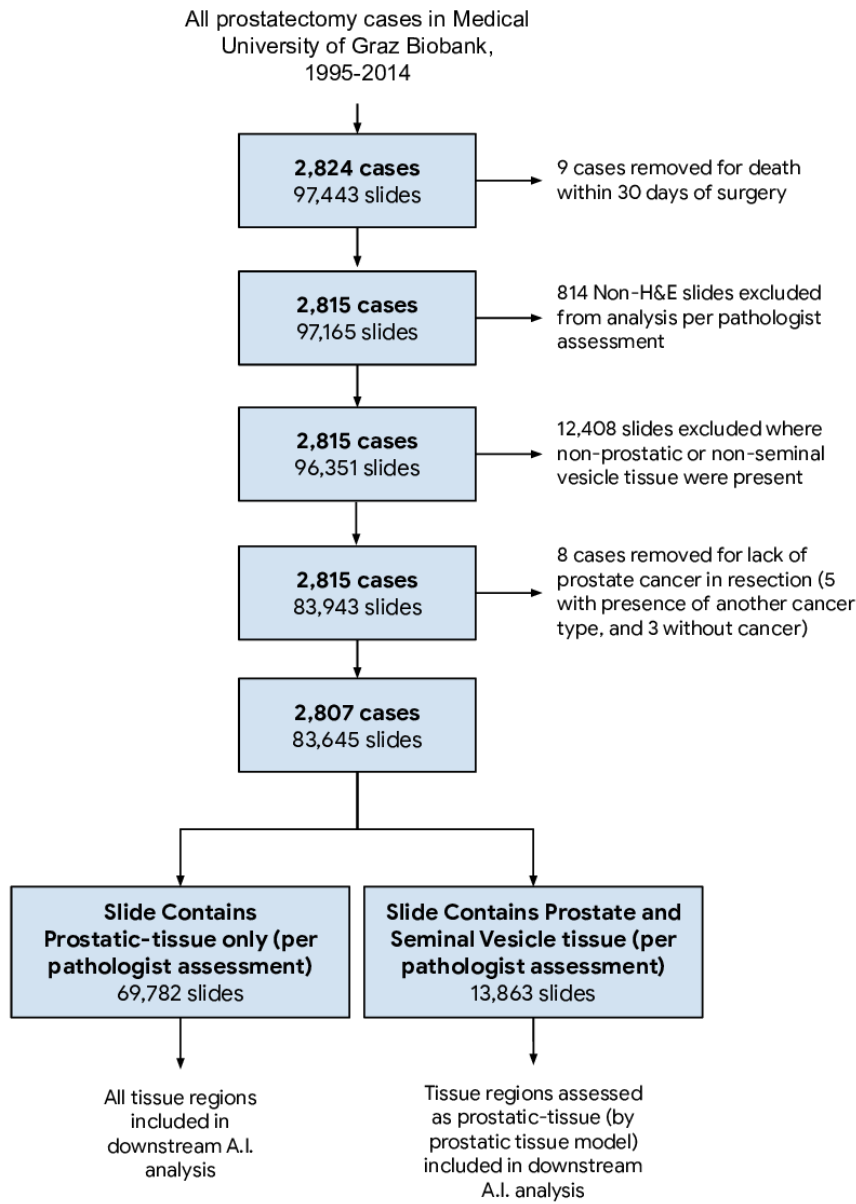

**Supplementary Figure S1. STARD diagram of inclusion/exclusion criteria.**

**(A)**

|  | Univariable C-index [95%CI] | | Multivariable C-index [95%CI] | |
| --- | --- | --- | --- | --- |
|  | Validation Set 1 | Validation Set 2 | Validation Set 1 | Validation Set 2 |
| Pathologist Grade Groups | N/A* | 0.79 [0.71-0.86] | N/A* | 0.87 [0.83-0.91] |
| A.I. risk score (continuous) | 0.83 [0.80-0.87] | 0.87 [0.81-0.91] | 0.85 [0.82-0.88] | 0.90 [0.85-0.94] |
| A.I. risk groups (discretized) | 0.82 [0.78-0.85] | 0.85 [0.79-0.90] | 0.83 [0.79-0.86] | 0.90 [0.86-0.93] |
| Average of A.I. Risk Groups and Pathologist Grade Groups (discretized) | N/A* | 0.86 [0.80-0.91] | N/A* | 0.89 [0.84-0.93] |

**(B) A.I. Risk Groups within Pathologic T-Stage 1-2 Cases**

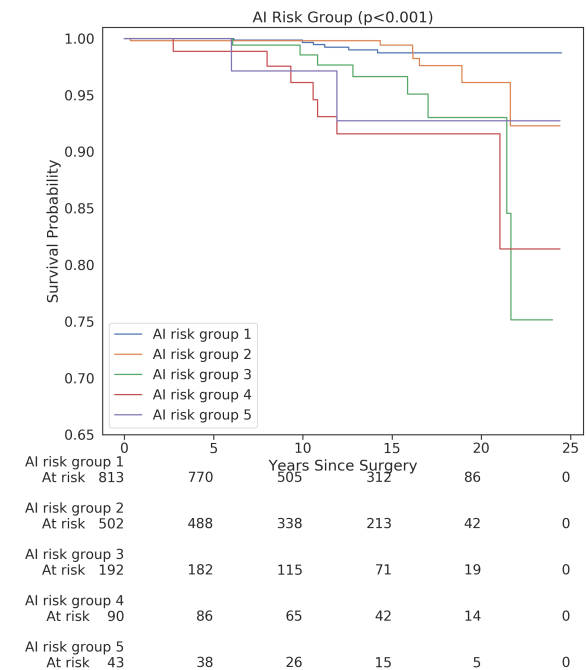

**(C) A.I. Risk Groups within Pathologic T-Stage 3-4 Cases:**

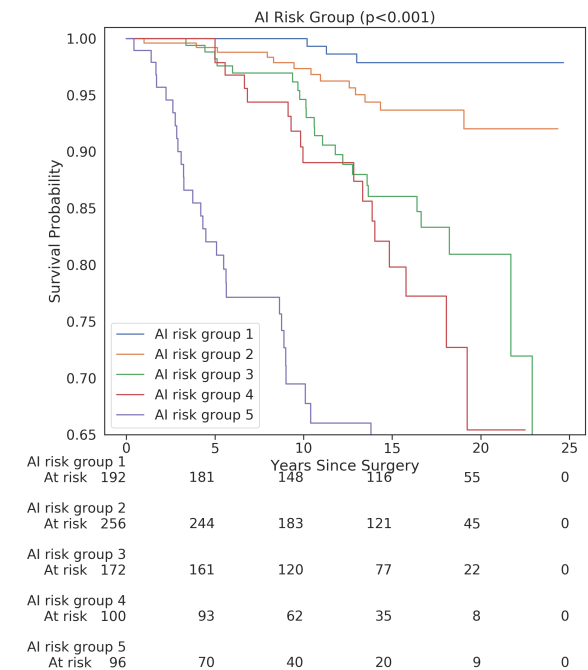

**Supplementary Figure S2. Multivariable and subgroup analyses involving pathologic T-stage.** (A) C-indices provided by univariable and multivariable Cox regression models, where multivariable regression utilized Grade Group and T-Stage. T-stage was categorized as 1-2, 3-4 (See Table 1), and multivariable Cox regression was fit with an L2 penalty of 0.02 to assist with convergence. (B-C) Kaplan-Meier curves for A.I. risk groups within pathologic T-stage categories of 1-2 and 3-4 for validation set 1